\documentclass[utf8]{frontiersSCNS} 

\usepackage{url,hyperref,lineno,microtype,subcaption}
\usepackage[onehalfspacing]{setspace}

\newcommand{\bh}{{\mathbf{h}}}
\newcommand{\bt}{{\mathbf{t}}}
\newcommand{\bv}{{\mathbf{v}}}
\newcommand{\bw}{{\mathbf{w}}}
\newcommand{\bg}{{\mathbf{g}}}

\newcommand{\bb}{{\mathbf{b}}}
\newcommand{\bmu}{{\mathbf{\mu}}}
\newcommand{\bsig}{{\mathbf{\sigma}}}
\newcommand{\bx}{{\mathbf{x}}}
\newcommand{\by}{{\mathbf{y}}}

\newcommand{\KL}{{\mathrm{KL}}}

\def\keyFont{\fontsize{8}{11}\helveticabold }
\def\firstAuthorLast{McClure {et~al.}} 
\def\Authors{Patrick McClure$^{1,2}$, Nao Rho$^{1,2}$, John A. Lee$^{2,3}$, Jakub R. Kaczmarzyk$^4$, Charles Zheng$^{1,2}$, Satrajit S. Ghosh$^4$, Dylan M. Nielson$^{2,3}$, Adam G. Thomas$^{3}$, Peter Bandettini$^2$, Francisco Pereira$^{1,2}$}
\def\Address{$^{1}$Machine Learning Team, National Institute of Mental Health, Bethesda, MD, USA \\
$^{2}$Section on Functional Imaging Methods, National Institute of Mental Health, Bethesda, MD, USA \\
$^{3}$Data Sharing and Science Team, National Institute of Mental Health, Bethesda, MD, USA  \\
$^{4}$McGovern Institute for Brain Research, Massachusetts Institute of Technology, Cambridge, MA, USA
}

\def\corrAuthor{Patrick McClure}

\def\corrEmail{patrick.mcclure@nih.gov}

\begin{document}
\onecolumn
\firstpage{1}

\title[Knowing what you know in brain segmentation]{Knowing what you know in brain segmentation using {Bayesian} deep neural networks} 

\author[\firstAuthorLast ]{\Authors} 
\address{} 
\correspondance{} 

\extraAuth{}

\maketitle

\begin{abstract}

\section{}

In this paper, we describe a Bayesian deep neural network (DNN) for predicting FreeSurfer segmentations of structural MRI volumes, in minutes rather than hours. The network was trained and evaluated on a large dataset (n = 11,480), obtained by combining data from more than a hundred different sites, and also evaluated on another completely held-out dataset (n = 418). The network was trained using a novel spike-and-slab dropout-based variational inference approach. We show that, on these datasets, the proposed Bayesian DNN outperforms previously proposed methods, in terms of the similarity between the segmentation predictions and the FreeSurfer labels, and the usefulness of the estimate uncertainty of these predictions. In particular, we demonstrated that the prediction uncertainty of this network at each voxel is a good indicator of whether the network has made an error and that the uncertainty across the whole brain can predict the manual quality control ratings of a scan. The proposed Bayesian DNN method should be applicable to any new network architecture for addressing the segmentation problem.


\tiny
 \keyFont{ \section{Keywords:} brain segmentation, deep learning, magnetic resonance imaging, bayesian neural networks, variational inference, automated quality control} 
\end{abstract}

\section{Introduction}

Identifying which voxels in a structural magnetic resonance imaging (sMRI) volume correspond to different brain structures (i.e. segmentation) is an essential processing step in neuroimaging analyses. These segmentations are often generated using the FreeSurfer package \citep{fischl2012FreeSurfer}, a process which can take a day or more for each subject \citep{Runtimes}. The computational resources for doing this at a scale of hundreds to thousands of subjects are beyond the capabilities of the computational resources available to most researchers. This has led to an interest in the use of deep neural networks as a general approach for learning to predict the outcome of a processing task, given the input data, in a much shorter time period than the processing would normally take. In particular, several deep neural networks have been trained to perform segmentation of brain sMRI volumes \citep{ronneberger2015u,roy2018quicknat,fedorov2017end,fedorov2017almost,li2017compactness,dolz20183d}, taking between a few seconds and a few minutes per volume. These networks predict a manual or an automated segmentation from the structural volumes (\cite{roy2018quicknat}, \cite{fedorov2017end}, \cite{fedorov2017almost}, and \cite{dolz20183d} used FreeSurfer, and \cite{petersen2010alzheimer} used GIF \citep{cardoso2015geodesic}).
%
These networks, however, have been trained on a limited number (on the order of hundreds) of examples from a limited number of sites (i.e. locations and/or scanners), which can lead to poor cross-site generalization for complex segmentation tasks with a large number of classes \citep{mcclure2018distributed}. This includes several of the recent DNNs proposed for fine-grain sMRI segmentation. (Note: We focus on DNNs which predict $>$30 classes.)

\cite{roy2018quicknat} performed 33 class segmentation using 581 sMRI volumes from the IXI dataset to train an initial model and then fine-tuned on 28 volumes from the MALC dataset \citep{marcus2007open}. They showed an approximately 9.4\% average Dice loss on out-of-site data from the ADNI-29 \citep{mueller2005alzheimer}, CANDI \citep{kennedy2012candishare}, and IBSR \citep{rohlfing2012image} datasets.
\cite{fedorov2017almost} used 770 sMRI volumes from HCP \citep{van2013wu} to train an initial model and then fine-tuned on 7 volumes from the FBIRN dataset \citep{keator2016function}. \cite{li2017compactness} performed a 160 class segmentation using 443 sMRI volumes from the ADNI dataset \citep{petersen2010alzheimer} for training. \cite{fedorov2017almost} and \cite{li2017compactness} did not report test results for sites that where not used during training.

These results show that it is possible to train a neural network to carry out segmentation of a sMRI volume. However, they provide a limited indication of whether such a network would work on data from any new site not encountered in training. While fine-tuning on labelled data from new sites can improve performance, even while using small amounts of data \citep{fedorov2017almost,roy2018quicknat,mcclure2018distributed}, a robust neural network segmentation tool should generalize to new sites without any further effort.
As part of the process of adding segmentation capabilities to the ``Nobrainer" tool \footnote{\url{https://github.com/neuronets/nobrainer}}, we trained a network to predict FreeSurfer segmentations given a training set of $\sim$10,000 sMRI volumes. This paper describes this process, as well as a quantitative and qualitative evaluation of the performance of the resulting model.


Beyond the segmentation performance of the network, a second aspect of interest to us is to understand whether it is feasible for a network to indicate how confident it is about its prediction at each location in the brain. We expect the network to make errors, be it because of noise, unusual positioning of the brain, very different contrast than what it was trained on, etc. Because our model is probabilistic and seeks to learn uncertainties, we expect it to be less confident in its predictions in such cases. It is also possible that, for certain locations, there are very similar brain structures labelled as different regions in different people. In such locations, there would be a limit to how well the network could perform, the Bayes error rate \citep{hastie2005elements}. Additionally, the network should be less confident for examples that are very different from those seen in the training set (e.g., contain large artifacts). While prediction uncertainty can be computed for standard neural networks, as done by  \cite{dolz20183d}, these uncertainty estimates are often overconfident \citep{guo2017calibration,mcclure2017representing}. Bayesian neural networks (BNNs) have been proposed as a solution to this issue. One popular BNN approach is Monte-Carlo (MC) Bernoulli Dropout \citep{srivastava2014dropout,gal2016dropout}. Using this method, \cite{li2017compactness,roy2018bayesian} showed that the segmentation performance of the BNN predictions was better for voxels with low dropout sampling-based uncertainties and that injected input noise can lead to increased uncertainty. \cite{roy2018bayesian} also found that using MC Bernoulli dropout decreased the drop in segmentation performance from 9.4\% to 7.8\% on average when testing on data from new sites compared to \cite{roy2018quicknat}. However, MC Bernoulli dropout does not learn dropout probabilities from data, which can lead to not properly modeling the uncertainty of the predicted segmentation. Recent works has shown that these dropout probabilities can be learned using a concrete relaxation \citep{gal2017concrete}. Additionally, learning individual uncertainties for each weight has been shown to be beneficial for many purposes (e.g. pruning and continual learning) \citep{blundell2015weight,nguyen2018variational,mcclure2018distributed}. In this paper, we propose using both learned dropout uncertainties and individual weight uncertainties.


Finally, we test the hypothesis that overall prediction uncertainty across an entire image reflects its ``quality", as measured by human quality control (QC) scores. Given the effort required to produce such scores, there have been multiple attempts to either crowdsource the process \citep{Keshavan363382}  or automate it \citep{esteban2017mriqc}. The latter, in particular, does not rely on segmentation information, so we believe it is worthwhile to test whether uncertainty derived from segmentation is more effective.


\section{Methods}

\begin{table} [h!]
\centering
\begin{tabular}{|c|c|c|}
\hline
{\bf Dataset} & {\bf Number of Examples}  \\
\hline
CoRR \citep{zuo2014open} & 3,039 \\
\hline
OpenfMRI \citep{poldrack2013toward} & 1,873 \\
\hline
NKI \citep{nooner2012nki} & 1,136 \\
\hline
SLIM \citep{liu2017longitudinal} & 1,003 \\
\hline
ABIDE \citep{di2014autism} & 992 \\
\hline
HCP \citep{van2013wu} & 956 \\
\hline
ADHD200 \citep{bellec2017neuro} & 719 \\
\hline
 CMI \citep{alexander2017open} & 611 \\
\hline
SALD \citep{wei2018structural} & 477 \\
\hline
Buckner \citep{biswal2010toward} & 183 \\
\hline
HBNSSI \citep{o2017healthy} & 178 \\
\hline
GSP \citep{holmes2015brain} & 152 \\
\hline
Haxby \citep{haxby2011common,nastase2017attention} & 55 \\
\hline
Gobbini \citep{di2017neural} & 51 \\
\hline
ICBM \citep{mazziotta2001probabilistic} & 45 \\
\hline
Barrios \citep{barrios} & 10 \\
\hline

\end{tabular}
\caption{The number of examples used from different datasets.}
\label{n}
\end{table}

\subsection{Data}

\subsubsection{Imaging Datasets}
We combined several datasets (Table \ref{n}), many of which themselves contain data from multiple sites, into a single dataset with 11,480 T1 sMRI volumes. In-site validation and test sets were created from the combined dataset using a 80-10-10 training-validation-test split. This resulted in a training set of 9,184 volumes, a validation set of 1,148 volumes, and a test set of 1,148 volumes. The training set was used for training the networks, the validation set for setting DNN hyperparameters (e.g, Bernoulli dropout probabilities), and the test set was used for evaluating the performance of the DNNs on new data from the same sites that were used for training.

We additionally used 418 sMRI volumes from the NNDSP dataset \citep{lee2018automated}  as a held-out dataset to test generalization of the network to an unseen site. In addition to sMRI volumes, each NNDSP sMRI volume was given a QC score from 1 to 4, higher scores corresponding to worse scan quality, by two raters (3 if values differed by more than 1), as described in \cite{blumenthal2002motion}. If a volume had a QC score greater than 2, it was labeled as a bad quality scan; otherwise, the scan was labeled as a good quality scan.

\subsubsection{Segmentation Target}

We computed 50-class FreeSurfer \citep{fischl2012FreeSurfer} segmentations, as in \cite{fedorov2017almost}, for all subjects in each of the datasets described earlier. These were used as the labels for prediction. Although, FreeSurfer segmentations may not be perfectly correct, as compared to manual, expert segmentations, using them allowed us to create a large training dataset, as one could not feasibly label it by hand.
FreeSurfer trained networks can also outperform FreeSurfer segmentations when compared to expert segmentations \citep{roy2018quicknat}.
The trained network could be fine-tuned with expert small amounts of labeled data, which would likely improve the results \citep{roy2018quicknat,mcclure2018distributed}.

\subsubsection{Data Pre-processing}

The sMRI volumes were resampled to 1mm isotropic cubic volumes of 256 voxels per side and the voxel intensities were normalized according to Freesurfer's mri\_convert with the conform flag. After resampling, input volumes were individually z-scored across voxels. We then split each sMRI volume into 512 non-overlapping $32\times32\times32$ sub-volumes, similarly to \citep{fedorov2017end,fedorov2017almost}, to be used as inputs for the neural network. The prediction target is the corresponding segmentation sub-volume. This resulted in 512 pairs, $(\bx, \by)$, of sMRI and label sub-volumes, respectively, for each sMRI volume.

\begin{table}
\centering
\begin{tabular}{|c|c|c|c|c|c|}
\hline
Layer & Filter & Padding & Dilation ($l$) & Non-linearity \\
\hline
1 & $96\textnormal{x}3^3$ & 1 & 1 & ReLU \\
\hline
2 & $96\textnormal{x}3^3$ & 1 & 1 & ReLU \\
\hline
3 & $96\textnormal{x}3^3$ & 1 & 1 & ReLU \\
\hline
4 & $96\textnormal{x}3^3$ & 2 & 2 & ReLU \\
\hline
5 & $96\textnormal{x}3^3$ & 4 & 4 & ReLU \\
\hline
6 & $96\textnormal{x}3^3$ & 8 & 8 & ReLU \\
\hline
7 & $96\textnormal{x}3^3$ & 1 & 1 & ReLU \\
\hline
8 & $50\textnormal{x}1^3$ & 0 & 1 & Softmax \\
\hline
\end{tabular}
\caption{The MeshNet dilated convolutional neural network architecture used for brain segmentation.}
\label{Arch}
\end{table}

\subsection{Convolutional Neural Network}

\subsubsection{Architecture}

Several deep neural network architectures have been proposed for brain segmentation, such as U-net \citep{ronneberger2015u}, QuickNAT \citep{roy2018quicknat}, HighResNet \citep{li2017compactness} and MeshNet \citep{fedorov2017end,fedorov2017almost}. We chose MeshNet because of its relatively simple structure, its lower number of learned parameters, and its competitive performance, since the computational cost of Bayesian neural networks scales based on structural complexity and number of parameters. 

MeshNet uses dilated convolutional layers \citep{yu2015multi} due to the 3D structural nature of sMRI data. Applying a discrete volumetric dilated convolutional layer to one input channel for one weight filter can be expressed as:

\begin{equation}
(\bw_f*_l\bh)_{i,j,k} = \sum_{\tilde{i}=-a}^a \sum_{\tilde{j}=-b}^b \sum_{\tilde{k}=-c}^c w_{f,\tilde{i},\tilde{j},\tilde{k}}h_{i-l\tilde{i},j-l\tilde{j},k-l\tilde{k}} = (\bw_f*_l\bh)_\bv = \sum_{\bt \in \mathcal{W}_{abc}} w_{f,\bt}h_{\bv-l\bt}.
\end{equation}

\noindent where $h$ is the input to the layer, $a$, $b$, and $c$ are the bounds for the $i$, $j$, and $k$ axes of the filter with weights $\bw_f$, $(i, j, k)$ is the voxel, $\bv$, where the convolution is computed. The set of indices for the elements of $\bw_f$ can be defined as $\mathcal{W}_{abc} = \{-a,...,a\}\times\{-b,...,b\}\times\{-c,...,c\}$. The dilation factor, number of filters, and other details of the MeshNet-like architecture that we used for all experiments is shown in Table \ref{Arch}. Note that we increased the number of filters per layer from 72 to 96, compared to \cite{fedorov2017almost} and \cite{mcclure2018distributed}, since we greatly increased the number of training volumes.

\subsubsection{Maximum a Posteriori Estimation}

When training a neural network, the weights of the network, $\bw$, are often learned using maximum likelihood estimation (MLE). For MLE, $\log p(\mathcal{D}|\bw)$ is maximized where $\mathcal{D} = \{(\bx_1, \by_1),...,(\bx_N, \by_N)\}$ is the training dataset and $(\bx_n, \by_n)$ is the $n$th input-output example. This often overfits, however, so we used a prior on the network weights, $p(\bw)$, to obtain a  maximum a posteriori (MAP) estimate, by optimizing $\log p(\bw|\mathcal{D})$:

\begin{equation}
\label{L_MAP}
\bw^* = \underset{\bw}{\mathrm{argmax}}\sum^N_{n=1} \log p(\by_n|\bx_n,\bw) + \log p(\bw).
\end{equation}

We used a fully factorized Gaussian prior (i.e. $p(w_{f,\tilde{i},\tilde{j},\tilde{k}}) = \mathcal{N}(0,1))$. This results in the MAP weights being learned by minimizing the softmax cross-entropy with L2 regularization. At test time, this point estimate approximation, $\bw^*$, is used to make a prediction for new examples:

\begin{equation}
p(\by_{test}|\bx_{test}) \approx p(\by_{test}|\bx_{test},\bw^*)  
\end{equation}

\begin{figure} [b]
    \centering
    \includegraphics[width=0.8\textwidth]{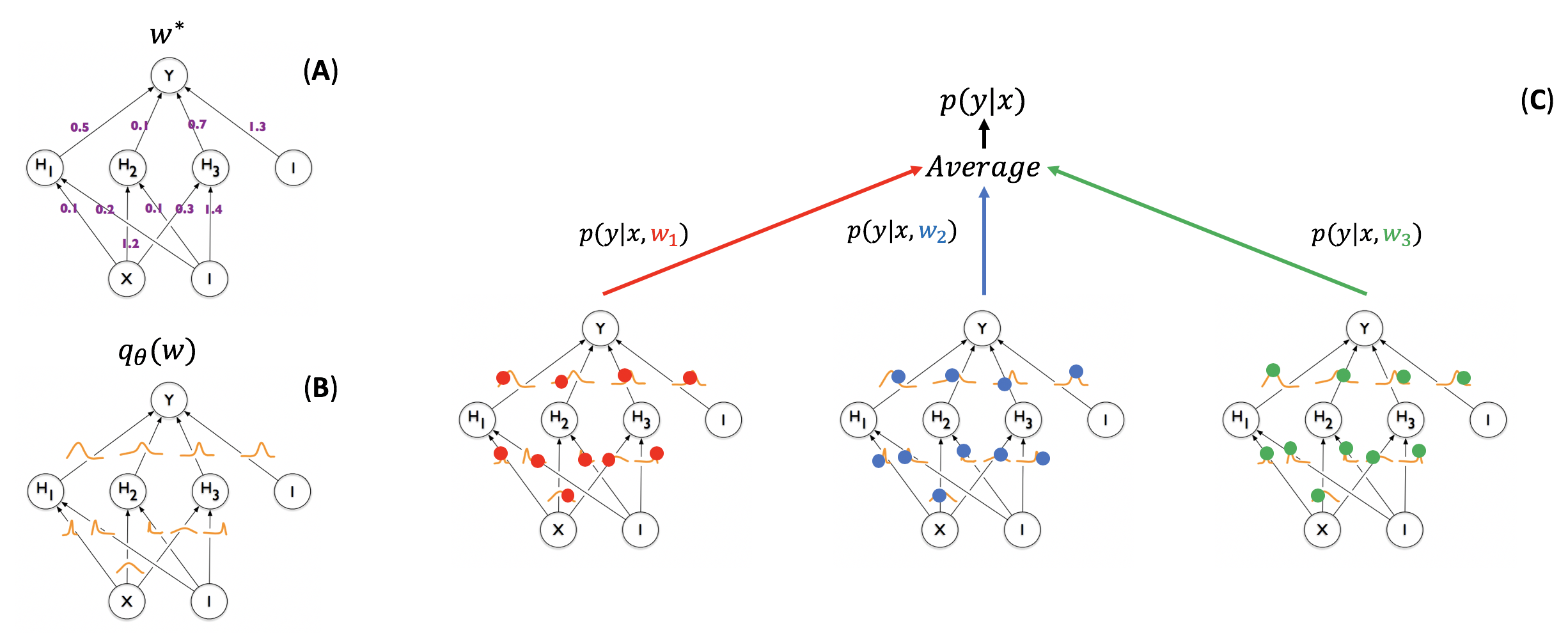}
    \caption{Illustration of generating a prediction from a Bayesian neural network using Monte Carlo sampling (modified from \cite{blundell2015weight}). A standard neural network ({\bf A}, top left) has one weight for each of its connections ($\bw^*$), learned from the training set and used in generating a prediction for a test example. A Bayesian neural network ({\bf B}, bottom left) has, instead, a posterior distribution for each weight, parameterized by theta ($q_{\theta}(\bw)$). The process of training starts with an assigned prior distribution for each weight, and returns an approximate posterior distribution. At test time ({\bf C}, right), a weight sample ${\bf w_1}$ (red) is drawn from the posterior distribution of the weights, and the resulting network is used to generate a prediction $p(y|x,{\bf w_1})$ for an example $x$. The same can be done for samples ${\bf w_2}$ (blue) and ${\bf w_3}$ (green), yielding predictions $p(y|x,{\bf w_2})$ and $p(y|x,{\bf w_3})$, respectively. The three networks are treated as an ensemble and their predictions averaged.}
    \label{fig:mc_sampling}
\end{figure}

\subsubsection{Approximate Bayesian Inference}

In Bayesian inference for neural networks, a distribution of possible weights is learned instead of just a MAP point estimate. Using Bayes' rule, $p(\bw|\mathcal{D})=p(\mathcal{D}|\bw)p(\bw)/p(\mathcal{D})$, where $p(\bw)$ is the prior over weights. However, directly computing the posterior, $p(\bw|\mathcal{D})$, is often intractable, particularly for DNNs. As a result, an approximate inference method must be used.

One of the most popular approximate inference methods for neural networks is variational inference, since it scales well to large DNNs. In variational inference, the posterior distribution $p(\bw|\mathcal{D})$ is approximated by a learned variational distribution of weights $q_\theta(\bw)$, with learnable parameters $\theta$. This approximation is enforced by minimizing the Kullback-Leibler divergence (KL) between $q_\theta(\bw)$, and the true posterior, $p(\bw|\mathcal{D})$, $\KL[q_\theta(\bw)||p(\bw|\mathcal{D})]$, which measures how $q_\theta(\bw)$ differs from $p(\bw|\mathcal{D})$ using relative entropy. This is equivalent to maximizing the variational lower bound \citep{hinton1993keeping,graves2011practical,blundell2015weight,kingma2015variational,gal2016dropout,molchanov2017variational,louizos2017multiplicative}, also known as the evidence lower bound (ELBO),

\begin{equation}
\label{L_ELBO}
\mathcal{L}_{ELBO}(\theta) = \mathcal{L}_{\mathcal{D}}(\theta) - \mathcal{L}_{KL}(\theta),
\end{equation}

where $\mathcal{L}_{\mathcal{D}}(\theta)$ is

\begin{equation}
\label{L_D}
\mathcal{L}_{\mathcal{D}}(\theta) = \sum^N_{n=1} \mathbb{E}_{q_\theta(\bw)}[\log p(\by_n|\bx_n,\bw)]
\end{equation}

and $\mathcal{L}_{KL}(\theta)$ is the KL divergence between the variational distribution of weights and the prior,

\begin{equation}
\label{L_KL}
\mathcal{L}_{KL}(\theta) = \KL[q_\theta(\bw)||p(\bw)]
\end{equation},

\noindent, which measures how $q_\theta(\bw)$ differs from $p(\bw)$ using relative entropy.

Maximizing $L_{\mathcal{D}}$ seeks to learn a $q_\theta(\bw)$ that explains the training data, while minimizing $L_{KL}$ (i.e. keeping $q_\theta(\bw)$ close to $p(\bw)$) prevents learning a $q_\theta(\bw)$ that overfits to the training data.

The objective function in Eq. \ref{L_ELBO} is usually impractical to compute for deep neural networks, due to both: (1) being a full-batch approach and (2) integrating over $q_\theta(\bw)$. (1) is often dealt with by using stochastic mini-batch optimization \citep{robbins1951stochastic} and (2) is often approximated using Monte Carlo sampling. As discussed in \cite{graves2011practical,kingma2015variational}, these methods can be used to perform stochastic gradient variational Bayes (SGVB) in deep neural networks. For each parameter update, an unbiased estimate of $\nabla_\theta \mathcal{L}_{\mathcal{D}}$ for a mini-batch, $\{(\bx_1,\by_1),...,(\bx_M,\by_M)\}$, is calculated using one weight sample, $\bw_m$, from $q_\theta(\bw)$ for each mini-batch example. This results in the following approximation to Eq. \ref{L_ELBO}:  

\begin{equation}
\mathcal{L}_{ELBO}(\theta) \approx \mathcal{L}_{\mathcal{D}}^{SGVB}(\theta) - \mathcal{L}_{KL}(\theta),
\end{equation}

where

\begin{equation}
\mathcal{L}_{\mathcal{D}}(\theta) \approx \mathcal{L}_{\mathcal{D}}^{SGVB}(\theta) = \frac{N}{M} \sum_{m = 1}^M \log p(\by_m|\bx_m,\bw_m).
\end{equation}

At test time, the weights, $\bw$ would ideally be marginalized out, $p(\by_{test}|\bx_{test}) = \int p(\by_{test}|\bx_{test},\bw) q_{\theta}(\bw) d\bw$, when making a prediction for a new example. However, this is often impractical to compute for DNNs, so a Monte-Carlo approximation is often used. This results in the prediction of a new example being made by averaging the predictions of multiple weight samples from $q_{\theta}(\bw)$ (Figure \ref{fig:mc_sampling}):

\begin{equation}
p(\by_{test}|\bx_{test}) \approx \frac{1}{N_{MC}}\sum_n^{N_{MC}}p(\by_{test}|\bx_{test},\bw_n)  
\end{equation}

\noindent where $\bw_n \sim q_{\theta}(\bw)$.

\paragraph{MC Bernoulli Dropout}
For MC Bernoulli dropout (BD) \citep{gal2016dropout}, we drew weights from $q_\theta(\bw)$ by drawing a Bernoulli random variable ($b_{i,j,k}\sim Bern(p_{l})$), where $i,j,k$ are the indices of the volume axes,  for every element of the layer, $l$, input, $\bf{h}$, and then elementwise multiplying $\bb$ and $\bh$ before applying the next dilated convolutional layer. This effectively sets the filter weights to zero when applied to a dropped element. \cite{gal2016dropout} approximated the KLD between this Bernoulli variational distribution and a zero-mean Gaussian by replacing the variational distribution with a mixture of Gaussians, resulting in an L2-like penalty. However, this can lead to pathological behaviour due to Bernoulli distributions not having support over all real numbers \citep{hron2018variational}. In Bernoulli dropout, $p_{l}$ codes for the uncertainty of the weights and is often set layerwise via hyperparameter search. (For our experiments, we found the best value of $p$ to be 0.9 after searching over the values of 0.95, 0.9, 0.75, and 0.5 using the validation set.) However, Bayesian models would ideally learn how uncertain to be for each weight.

\paragraph{Spike-and-Slab Dropout with Learned Model Uncertainty}
We propose a form of dropout that both learns the dropout probability for each filter using a concrete relaxation of dropout \citep{gal2017concrete}, and an individual uncertainty for each weight using fully factorized Gaussian (FFG) filters \citep{graves2011practical,blundell2015weight,molchanov2017variational,nguyen2018variational,mcclure2018distributed}. This is in contrast to previous spike-and-slab dropout methods, which did not learn the model (or epistemic) uncertainty \citep{der2009aleatory,kendall2017uncertainties} from data either by learning the dropout probabilities or by learning the variance parameter of the Gaussian components of the weights \citep{mcclure2017representing}. In our proposed method, we assume each of the $F$ filters are independent (i.e. $p(\bw) = \prod_{f=1}^F p(\bw_f)$), as done in previous FFG methods \citep{graves2011practical,blundell2015weight,molchanov2017variational,nguyen2018variational,mcclure2018distributed}. We then decompose each filter into a dropout-based component, $b_f$, and a Gaussian component, $\bg_f$, such that $\bw_f = b_f \bg_f$. Per this decomposition, we perform variational inference on the joint distribution of $\{b_1,...,b_F,\bg_1,...\bg_F\}$, instead of on p($\bw$) directly \citep{titsias2011spike,mcclure2017representing}. We then assume each element of $\bg_f$ is independent (i.e. $p(\bg_f) = \prod_{\bt \in \mathcal{W}_{abc}} p(g_{f,\bt})$), and that each weight is Gaussian (i.e. $g_{f,\bt} \sim \mathcal{N}(\mu_{f,\bt},\sigma_{f,\bt}^2)$) with learned parameters $\mu_{f,\bt}$ and $\sigma_{f,\bt}$. Instead of drawing each $b_{f}$ from $Bern(p_{l})$, we draw them from a concrete distribution \citep{gal2017concrete} with a learned dropout probability, $p_{f}$, for each filter:

\begin{equation}
    b_{f}=sigmoid\big( \frac{1}{t} (\log p_{f} - \log(1 - p_{f}) + \log u - \log(1 - u))
\end{equation}

\noindent where $u \sim Unif(0,1)$. This concrete distribution converges to the Bernoulli distribution as the sigmoid scaling parameter, $t$, goes to zero. (In this paper, we used $t=0.02$.) As discussed in \cite{kingma2015variational} and \cite{molchanov2017variational}, randomly sampling each $g_{f,\bt}$  for each mini-batch example can be computationally expensive, so we used the fact that the sum of independent Gaussian variables is also Gaussian to move the noise from the weights to the convolution operation, as in \cite{mcclure2018distributed}. For, dilated convolutions and the proposed spike-and-slab variational distribution, this is described by:

\begin{equation}
(\bw_f*_l \bh)_\bv = b_f (\bg_f *_l \bh)_\bv
\end{equation}

where

\begin{equation}
(\bg_f *_l \bh)_\bv \sim \mathcal{N}(\mu_{f,\bv}^*,(\sigma_{f,\bv}^*)^2),
\label{eq:w_f}
\end{equation}

\begin{equation}
\mu_{f,\bv}^* = \sum_{\bt \in \mathcal{W}_{abc}}\mu_{f,\bt}h_{\bv-l\bt},
\end{equation}

and

\begin{equation}
(\sigma_{f,\bv}^*)^2 = \sum_{\bt \in \mathcal{W}_{abc}}\sigma_{f,\bt}^2h_{\bv-l\bt}^2.
\end{equation}

For this spike-and-slab dropout (SSD) implementation, we used a spike-and-slab prior, instead of the Gaussian prior used by \cite{gal2016dropout} and \cite{gal2017concrete}. Using a spike-and-slab prior with MC Bernoulli dropout was discussed in \cite{gal2016uncertainty}, but not implemented. As in the variational distribution, each filter is independent in the prior. Per the spike-and-slab decomposition discussed above, the KL-divergence term of the ELBO can be written as

\begin{equation}
\mathcal{L}_{KL}(\theta) = \sum_{f=1}^F \KL[q_{p_f}(b_f)q_{\bmu,\bsig}(\bg_f)||p(b_f)p(\bg_f)], 
\end{equation}

 \noindent where $\theta = \bigcup_f^F \bigcup_{\bt \in \mathcal{W}_{abc}} \{p_f,\mu_{f,\bt},\sigma_{f,\bt}\}$ are the learned parameters and $p(b_f)$ and $p(\bg_f)$ are priors. Assuming that each weight in a filter is independent, as commonly done in the literature \citep{graves2011practical,blundell2015weight,nguyen2018variational}, allows the term to be rewritten as
 
 \begin{equation}
  \mathcal{L}_{KL}(\theta) = \sum_{f=1}^F (\KL[q_{p_f}||p(b_f)] + \sum_{\bt \in \mathcal{W}_{abc}} \KL[q_{\bmu,\bsig}(g_{f,\bt})||p(g_{f,\bt})]) .  
 \end{equation}
 
\noindent For $\KL[q_{p_f}||p(b_f)]$, we used the KL-divergence between two Bernoulli distributions, 

\begin{equation}
\KL[q_{p_f}(b_f)||p(b_f)] = p_f \log\frac{p_f}{p_{prior}} + (1-p_f) \log\frac{1-p_f}{1-p_{prior}}, 
\end{equation}

\noindent since we used a relatively small sigmoid scaling parameter. Using $p(g_{f,\bt}) = \mathcal{N}(\mu_{prior},\sigma_{prior}^2)$,

\begin{equation}
\KL[q_{\bmu,\bsig}(g_{f,\bt})||p(g_{f,\bt})] = \log\frac{\sigma_{prior}}{\sigma_{f,\bt}} + \frac{\sigma_{f,\bt}^2 + (\mu_{f,\bt}-\mu_{prior})^2}{2\sigma_{prior}^2} - \frac{1}{2}.   \end{equation}

 \noindent For this paper, the spike-and-slab prior parameters were set as $p_{prior} = 0.5$, $\mu_{prior} = 0$, and $\sigma_{prior} = 0.1$. $p_{prior} = 0.5$ corresponds to a maximum entropy prior (i.e. in the absence of new data be maximally uncertain). Alternatively, a $p_{prior}$ close to $0$ is a sparcity prior (i.e. in the absence of data do not use a filter).

\subsection{Implementation Details}
The DNNs were implemented using Tensorflow \citep{abadi2016tensorflow}. During training, the parameters of each DNN were updated using Adam \citep{kingma2014adam} with an initial learning rate of 1e-4. A mini-batch size of 32 subvolumes was used with data parallelization across 4 12GB NVIDIA Titan X Pascal GPUs was used for training and a mini-batch size of 8 subvolumes on 1 12GB NVIDIA Titan X Pascal GPU was used for validation and testing.

\subsection{Quantifying performance}
\subsubsection{Segmentation performance measure}

To measure the quality of the produced segmentations, we calculated the Dice coefficient, which is defined by

\begin{equation}
Dice_c = \frac{2|\hat{\by}_c \cdot \by_c|}{||\hat{\by}_c||^2 + ||\by_c||^2} = \frac{2TP_c}{2TP_c + FN_c + FP_c},\end{equation}

\noindent where $\hat{\by}_c$ is the binary segmentation for class $c$ produced by a network, $\by_c$ is the ground truth produced by FreeSurfer, $TP_c$ is the true positive rate for class $c$, $FN_c$ is the false negative rate for class $c$, and $FP_c$ is the false positive rate for class $c$. We calculate the Dice coefficient separately for each class $c=1,\ldots,50$, and average across classes to compute the overall performance of a network for one sMRI volume.

\subsubsection{Uncertainty measure}

We quantify the uncertainty of a prediction, $p(\by_{m,c}|x_m)$, using the aleatoric uncertainty \citep{der2009aleatory,kendall2017uncertainties}, which was measured by the entropy of the softmax across the 50 output classes,

\begin{equation}
H(\by_{m,c}|x_m) = -\sum_{c=1}^{50} p(\by_{m,c}|x_m) \ \log p(\by_{m,c}|x_m).
\end{equation}
\noindent We calculate the uncertainty for each output voxel separately, and the uncertainty for one sMRI volume by averaging across all output voxels not classified as background (i.e. given the unknown label).

\begin{table} [h!]
\centering
\begin{tabular}{|c|c|c|}
\hline
Method & In-site & Out-of-site  \\
\hline
MAP & 0.7790 $\pm$ 0.0576 & 0.7333 $\pm$ 0.0498 \\
\hline
BD & 0.7764 $\pm$ 0.0506 & 0.7369 $\pm$ 0.0474 \\
\hline
SSD & 0.8373 $\pm$ 0.0471 & 0.7921 $\pm$ 0.0444 \\
\hline

\end{tabular}
\caption{The average and standard deviation of the class Dices across test volumes for the maximum a posteriori (MAP), MC Bernoulli dropout (BD), and spike-and-slab dropout (SSD) network on the in-site and out-of-site test sets.}
\label{dices}
\end{table}

\section{Results}
\subsection{Segmentation performance}

\begin{figure}
    \centering
    \includegraphics[width=0.9\textwidth]{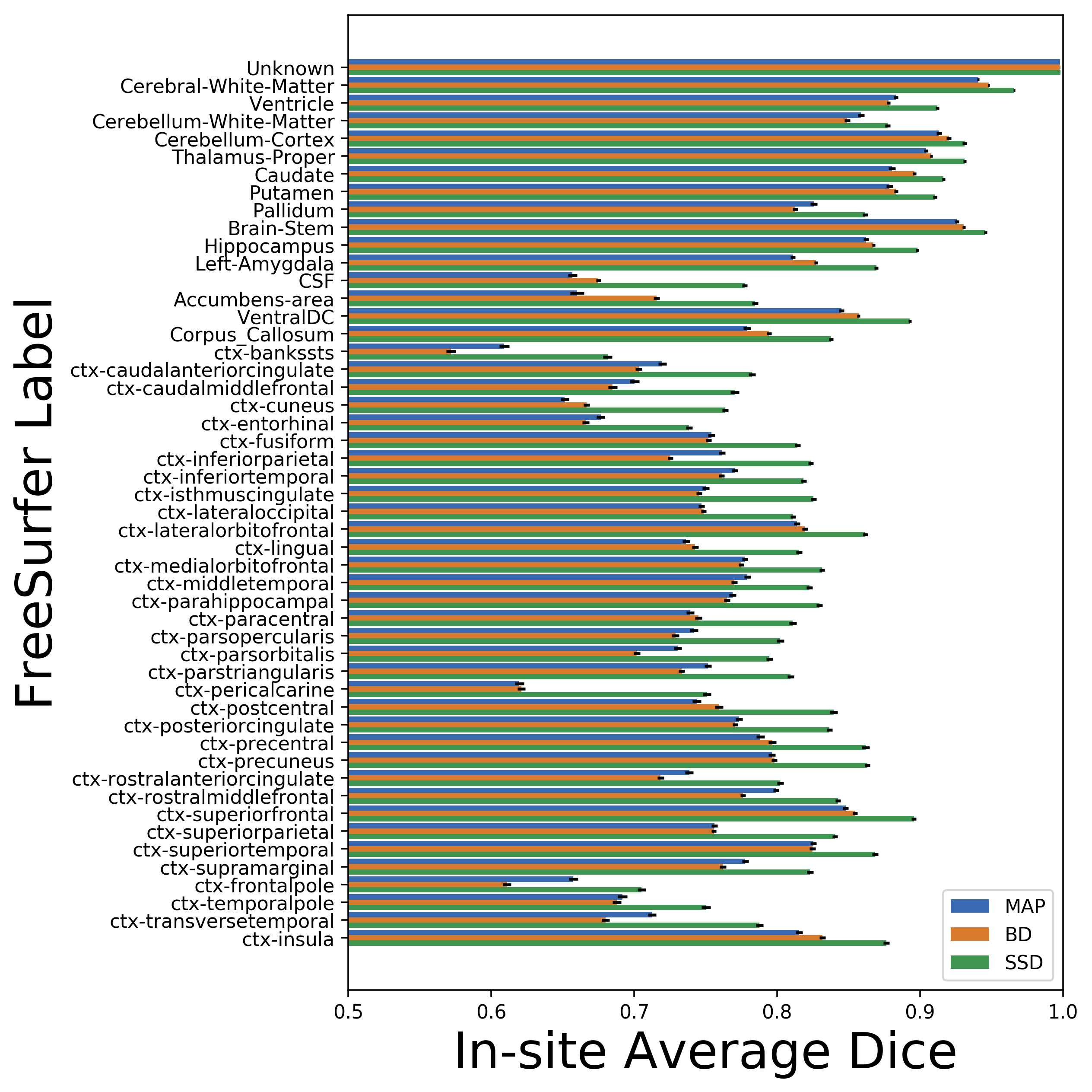}
    \caption{Average Dice scores and standard errors across in-site test volumes for each label for the maximum a posteriori (MAP), MC Bernoulli dropout (BD), and spike-and-slab dropout (SSD) networks.}\label{dices_in}
\end{figure}

\begin{figure}
    \centering
    \includegraphics[width=0.9\textwidth]{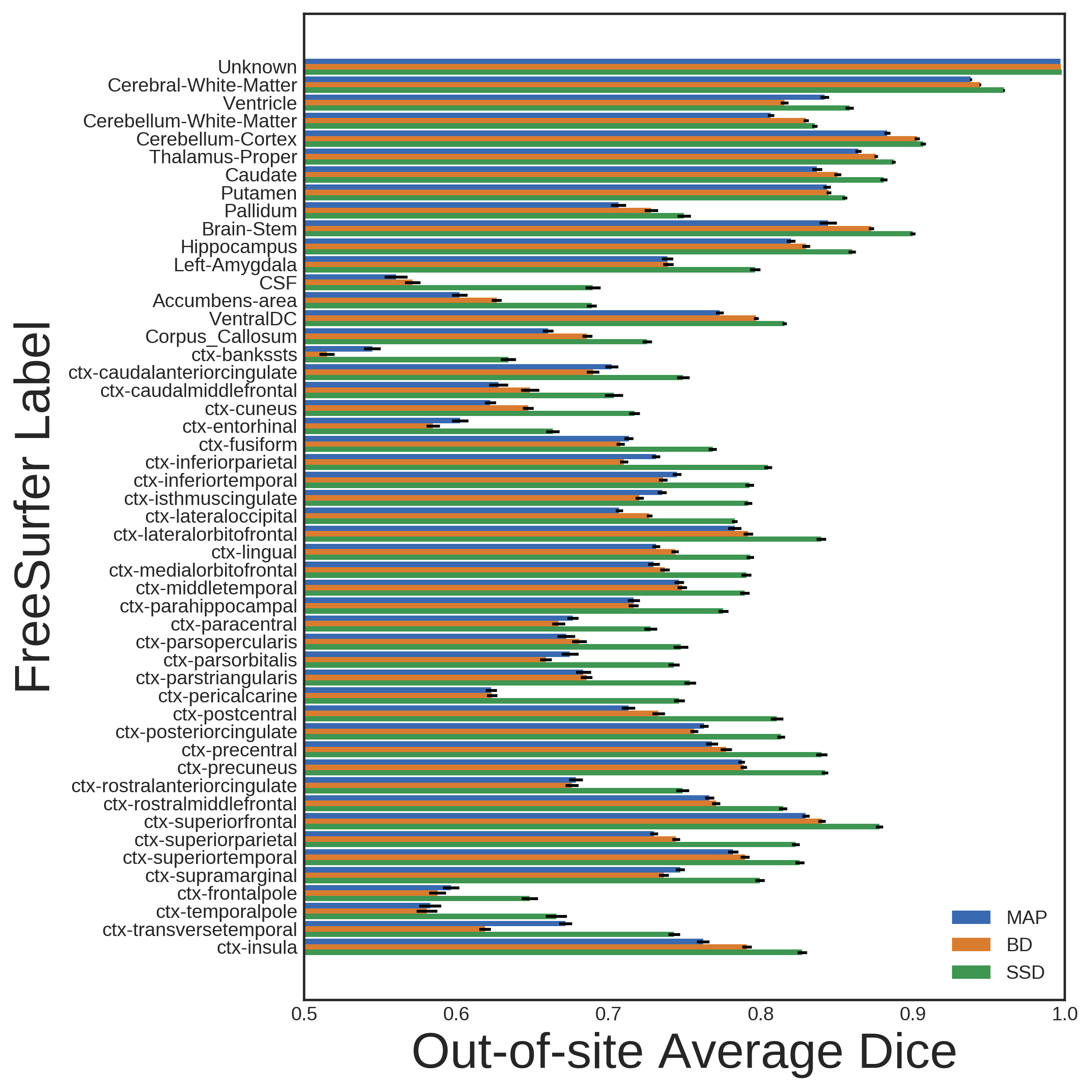}
    \caption{Average Dice scores and standard errors across out-of-site test volumes for each label for the maximum a posteriori (MAP), MC Bernoulli dropout (BD), and spike-and-slab dropout (SSD) networks.}\label{dices_out}
\end{figure}

\begin{figure}[!h]

\begin{subfigure}[b]{0.16\textwidth}
\includegraphics[scale=0.25]{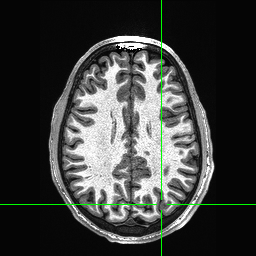}
\end{subfigure}
~
\begin{subfigure}[b]{0.16\textwidth}
\includegraphics[scale=0.25]{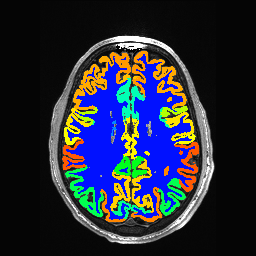}
\end{subfigure}
~
\centering 
\begin{subfigure}[b]{0.16\textwidth}
\includegraphics[scale=0.25]{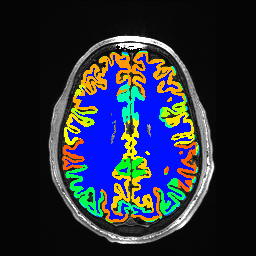}
\end{subfigure}
~
\begin{subfigure}[b]{0.16\textwidth}
\includegraphics[scale=0.25]{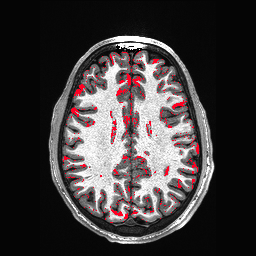}
\end{subfigure}
~
\begin{subfigure}[b]{0.16\textwidth}
\includegraphics[scale=0.25]{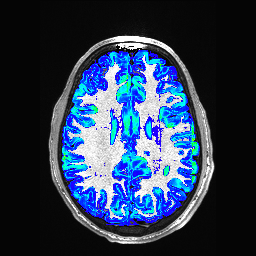}
\end{subfigure}
~
\begin{subfigure}[b]{0.05\textwidth}
\includegraphics[scale=0.072]{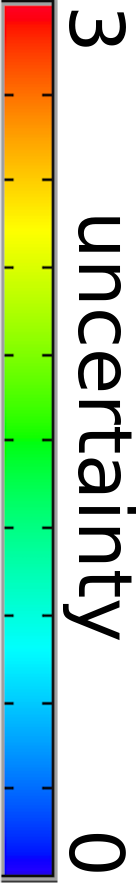}
\end{subfigure}


\begin{subfigure}[b]{0.16\textwidth}
\includegraphics[scale=0.25]{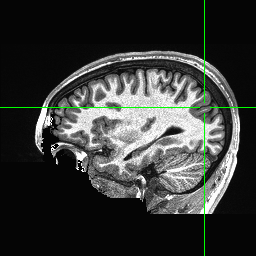}
\end{subfigure}
~
\begin{subfigure}[b]{0.16\textwidth}
\includegraphics[scale=0.25]{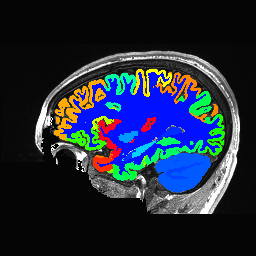}
\end{subfigure}
~
\centering 
\begin{subfigure}[b]{0.16\textwidth}
\includegraphics[scale=0.25]{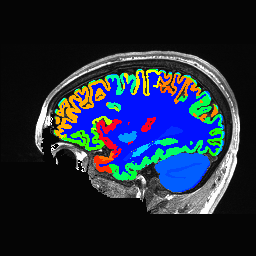}
\end{subfigure}
~
\begin{subfigure}[b]{0.16\textwidth}
\includegraphics[scale=0.25]{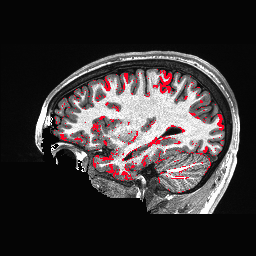}
\end{subfigure}
~
\begin{subfigure}[b]{0.16\textwidth}
\includegraphics[scale=0.25]{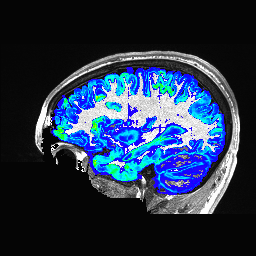}
\end{subfigure}
~
\begin{subfigure}[b]{0.05\textwidth}
\includegraphics[scale=0.072]{colorbar.png}
\end{subfigure} 


\begin{subfigure}[b]{0.16\textwidth}
\includegraphics[scale=0.25]{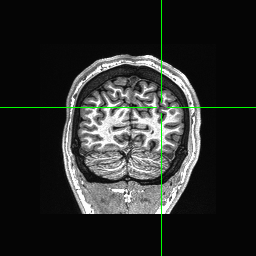}
\caption*{structural}
\end{subfigure}
~
\begin{subfigure}[b]{0.16\textwidth}
\includegraphics[scale=0.25]{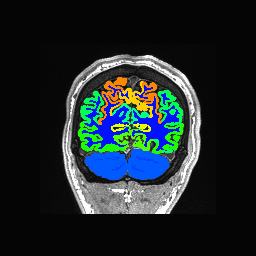}
\caption*{FreeSurfer}
\end{subfigure}
~
\centering 
\begin{subfigure}[b]{0.16\textwidth}
\includegraphics[scale=0.25]{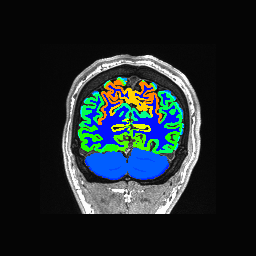}
\caption*{prediction}
\end{subfigure}
~
\begin{subfigure}[b]{0.16\textwidth}
\includegraphics[scale=0.25]{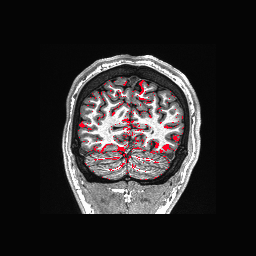}
\caption*{error}
\end{subfigure}
~
\begin{subfigure}[b]{0.16\textwidth}
\includegraphics[scale=0.25]{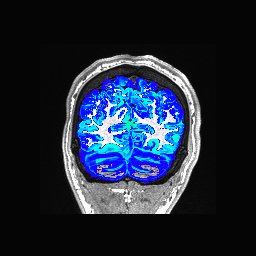}
\caption*{uncertainty}
\end{subfigure} 
~
\begin{subfigure}[b]{0.05\textwidth}
\includegraphics[scale=0.072]{colorbar.png}
\caption*{}
\end{subfigure} 

\caption{In-site segmentation results for the spike-and-slab dropout (SSD) network for a test subject with average Dice performance. The columns show, respectively, the structural image used as input, the FreeSurfer segmentation used as a prediction target, the prediction made by our network, the voxels where there was a mismatch between prediction and target, and the prediction uncertainty at each voxel.}
\label{fig:test}
\end{figure}

\begin{figure}[!h]


\begin{subfigure}[b]{0.16\textwidth}
\includegraphics[scale=0.25]{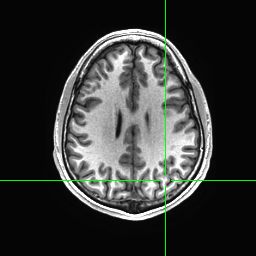}
\end{subfigure}
~
\begin{subfigure}[b]{0.16\textwidth}
\includegraphics[scale=0.25]{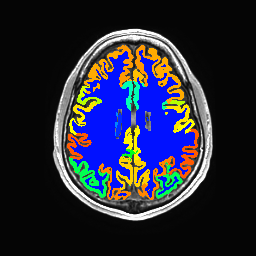}
\end{subfigure}
~
\centering 
\begin{subfigure}[b]{0.16\textwidth}
\includegraphics[scale=0.25]{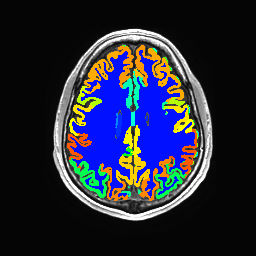}
\end{subfigure}
~
\begin{subfigure}[b]{0.16\textwidth}
\includegraphics[scale=0.25]{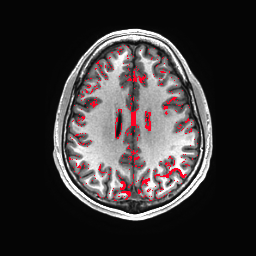}
\end{subfigure}
~
\begin{subfigure}[b]{0.16\textwidth}
\includegraphics[scale=0.25]{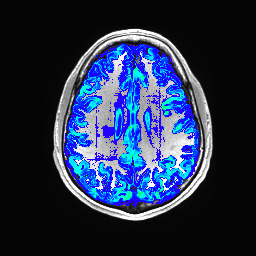}
\end{subfigure} 
~
\begin{subfigure}[b]{0.05\textwidth}
\includegraphics[scale=0.072]{colorbar.png}
\end{subfigure} 


\begin{subfigure}[b]{0.16\textwidth}
\includegraphics[scale=0.25]{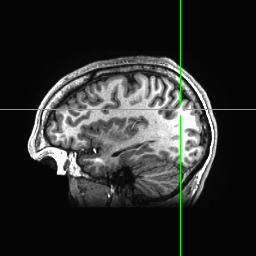}
\end{subfigure}
~
\begin{subfigure}[b]{0.16\textwidth}
\includegraphics[scale=0.25]{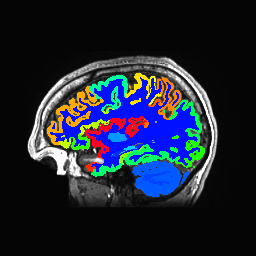}
\end{subfigure}
~
\centering 
\begin{subfigure}[b]{0.16\textwidth}
\includegraphics[scale=0.25]{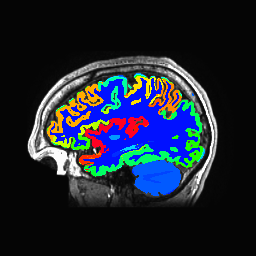}
\end{subfigure}
~
\begin{subfigure}[b]{0.16\textwidth}
\includegraphics[scale=0.25]{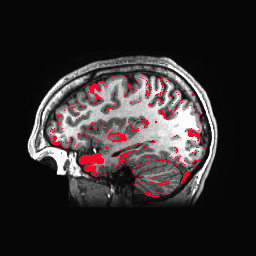}
\end{subfigure}
~
\begin{subfigure}[b]{0.16\textwidth}
\includegraphics[scale=0.25]{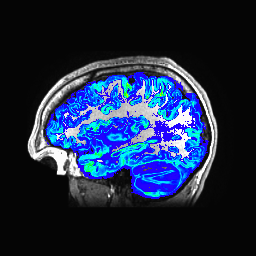}
\end{subfigure}
~
\begin{subfigure}[b]{0.05\textwidth}
\includegraphics[scale=0.072]{colorbar.png}
\end{subfigure} 


\begin{subfigure}[b]{0.16\textwidth}
\includegraphics[scale=0.25]{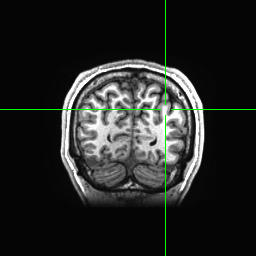}
\caption*{structural}
\end{subfigure}
~
\begin{subfigure}[b]{0.16\textwidth}
\includegraphics[scale=0.25]{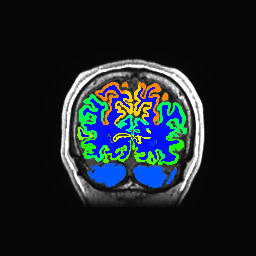}
\caption*{FreeSurfer}
\end{subfigure}
~
\centering 
\begin{subfigure}[b]{0.16\textwidth}
\includegraphics[scale=0.25]{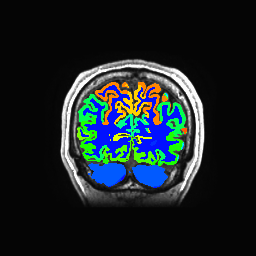}
\caption*{prediction}
\end{subfigure}
~
\begin{subfigure}[b]{0.16\textwidth}
\includegraphics[scale=0.25]{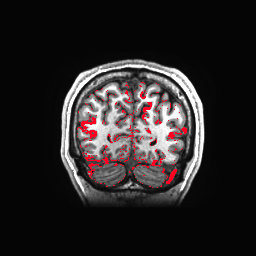}
\caption*{error}
\end{subfigure}
~
\begin{subfigure}[b]{0.16\textwidth}
\includegraphics[scale=0.25]{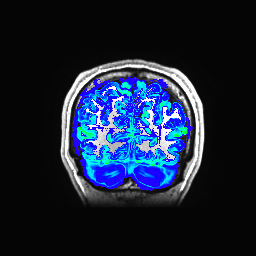}
\caption*{uncertainty}
\end{subfigure} 
~
\begin{subfigure}[b]{0.05\textwidth}
\includegraphics[scale=0.072]{colorbar.png}
\caption*{}
\end{subfigure} 

\caption{Out-of-site segmentation results for the spike-and-slab dropout (SSD) network for a test subject with average Dice performance. The columns show, respectively, the structural image used as input, the FreeSurfer segmentation used as a prediction target, the prediction made by our network, the voxels where there was a mismatch between prediction and target, and the prediction uncertainty at each voxel.}
\label{fig:nndsp}
\end{figure}

We trained MAP, MC Bernoulli Dropout (BD), and Spike-and-Slab Dropout (SSD) Meshnet-like CNNs on the 9,298 sMRI volumes in the training set. We then applied our networks to produce segmentations for both the in-site test set and the out-of-site test data. For the BD and SSD networks, 10 MC samples were used for test predictions. The means and standard deviations across volumes for the average Dice across all 50 classes are shown in Table \ref{dices}. Dice scores for each label for the in-site and out-of-site test sets are shown in Figure \ref{dices_in} and \ref{dices_out}, respectively. We found that, compared to MAP and BD, SSD significantly increased the Dice for both the in-site  ($p<1e-6$) and out-of-site ($p<1e-6$) test sets, per a paired t-test across test volumes. We found that SSD had a 5.7\% drop in performance from the in-site test set to the out-of-site test set, where as the MAP has a drop of 6.2\% and BD a drop of 5.4\%. This is better than drops of 9.4\% and 7.8\% on average reported in the literature by \cite{roy2018quicknat} and \cite{roy2018bayesian}, respectively. In Figures \ref{fig:test} and \ref{fig:nndsp}, we show selected example segmentations for the SSD network for volumes that have Dice scores similar to the average Dice score across the respective dataset.

\subsection{Utilizing Uncertainty}

\subsubsection{Predicting segmentation errors from uncertainty}
\begin{figure}
    \centering
    \includegraphics[width=0.4\textwidth]{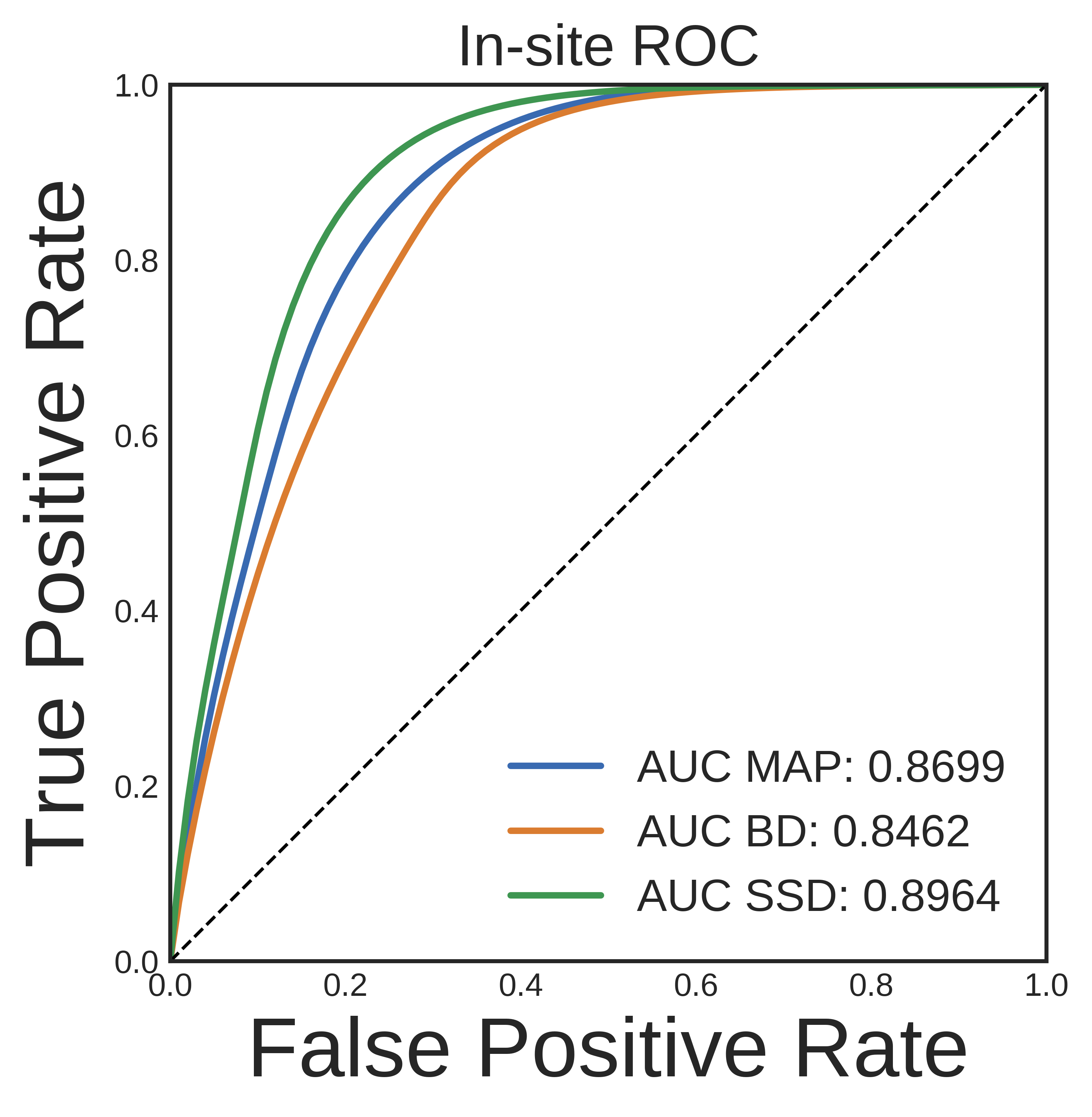}
    \includegraphics[width=0.4\textwidth]{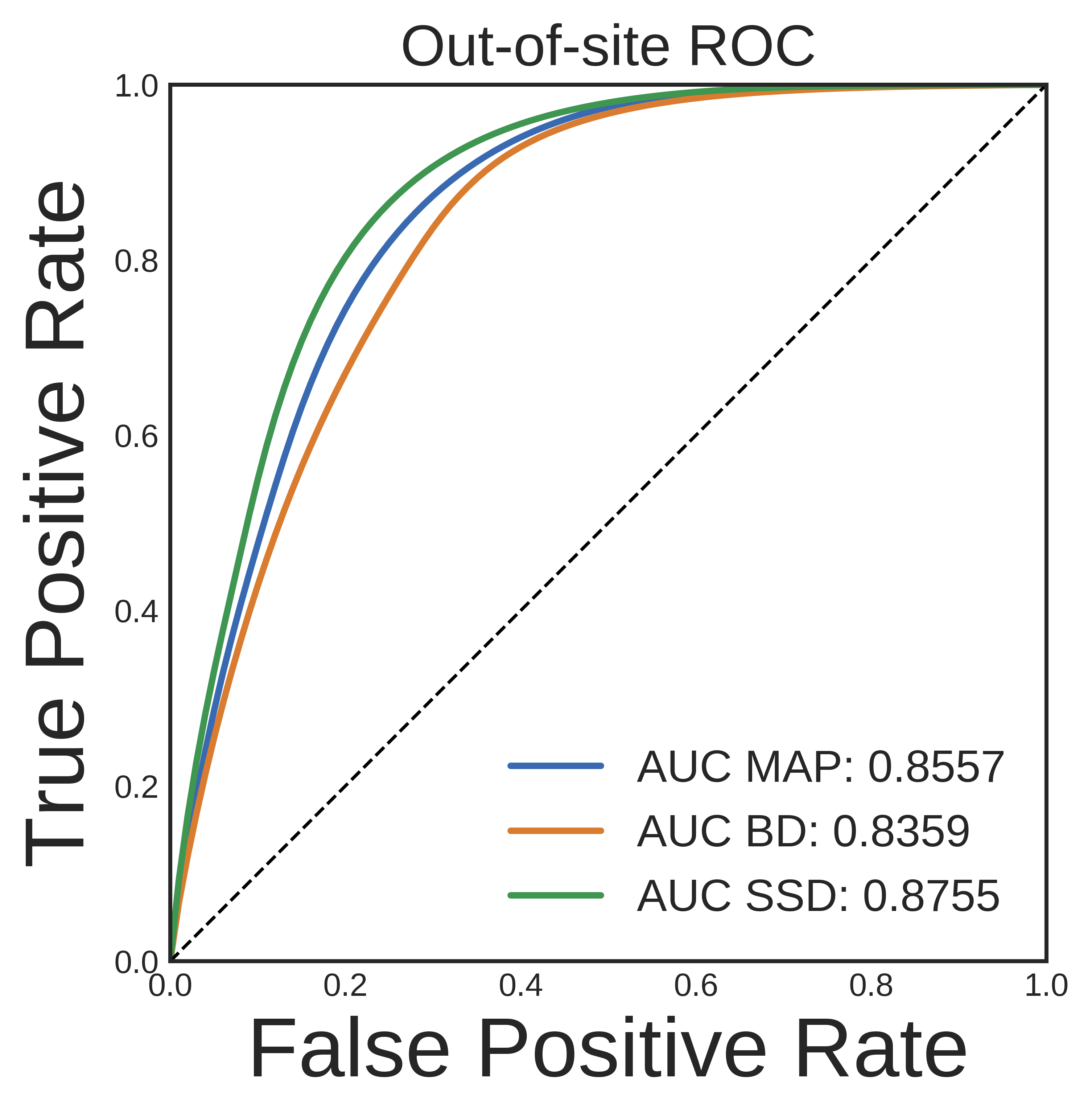}
    \caption{Receiver operating characteristic (ROC) curves for predicting errors for the in-site and out-of-site test sets from the voxel uncertainty of the maximum a posteriori (MAP), MC Bernoulli dropout (BD), and spike-and-slab dropout (SSD) networks.}
    \label{entropy_to_error}
\end{figure}

Ideally, an increase in DNN prediction uncertainty indicates an increase in the probability that that prediction is incorrect. To evaluate whether this is the case for the trained brain segmentation DNN, we performed a receiver operating characteristic (ROC) analysis. In this analysis, voxels are ranked from most uncertain to least uncertain and one considers, at each rank, what fraction of the voxels were also misclassified by the network. An ROC curve can then be generated by plotting the true positive rate vs the false negative rate for different uncertainty thresholds used to predict misclassification. The area under this curve (AUC) typically summarizes the results of the ROC analysis. The average ROC and AUCs across volumes for MAP, BD, and SSD for the in-site and out-of-site test sets are shown in \ref{entropy_to_error}. Compared to MAP and BD, SSD significantly improved the AUC for both the in-site ($p<1e-6$) and out-of-site ($p<1e-6$) test sets, per a paired t-test across test set volumes.

\subsubsection{Predicting scan quality from uncertainty}
\begin{figure} [!h]
    \centering
    \includegraphics[width=0.5\textwidth]{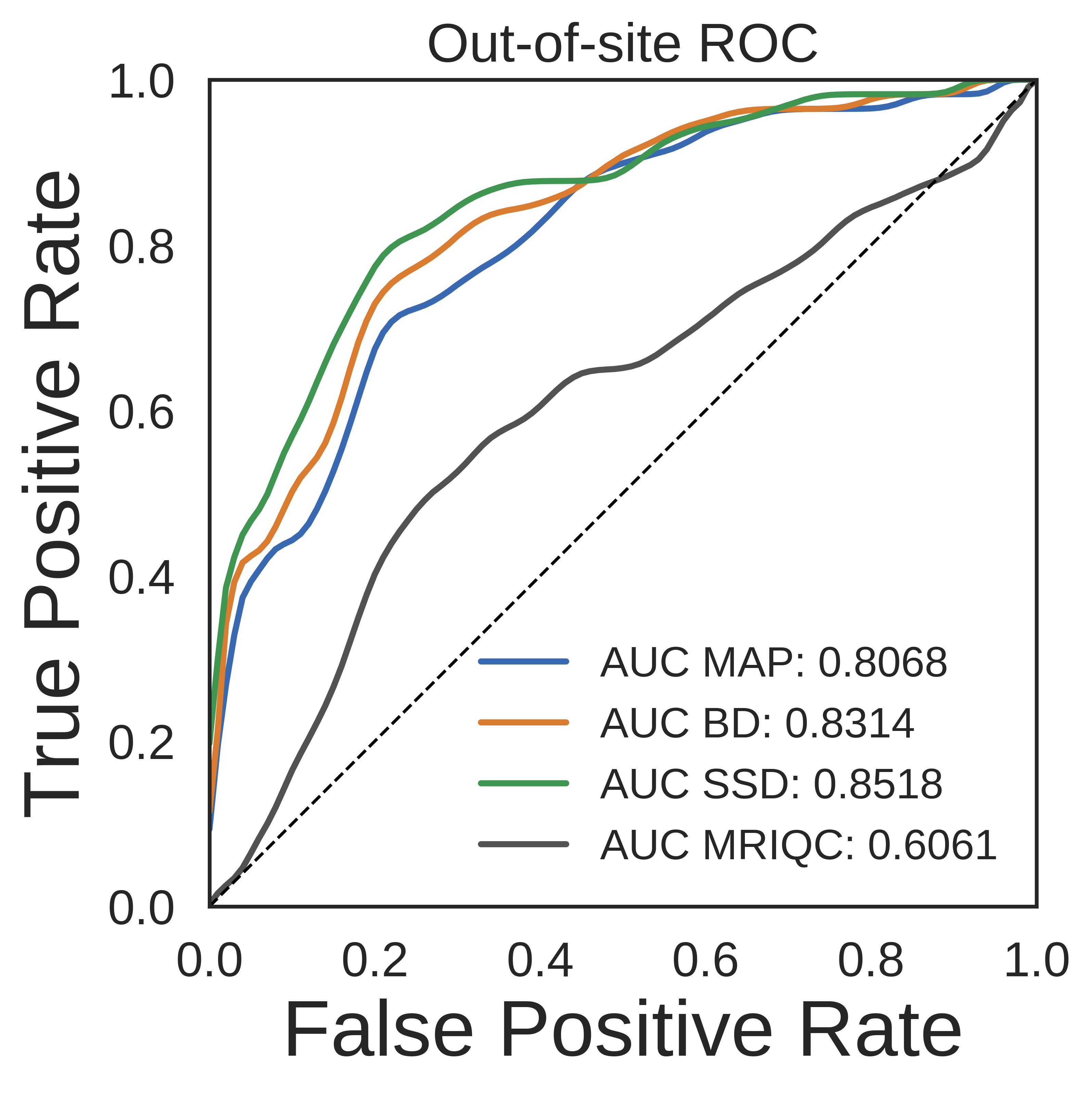}
    \caption{Receiver operating characteristic (ROC) curves for predicting scan quality for the NNDSP out-of-site test set from the average non-background voxel uncertainty of the maximum a posteriori (MAP), MC Bernoulli dropout (BD), and spike-and-slab dropout (SSD) networks and from MRIQC scores.}\label{qc_roc}
\end{figure}

Ideally, the output uncertainty for inputs not drawn from the training distribution should be relatively high. This could potentially be useful for a variety of applications. One particular application is detection of bad quality sMRI scans, since the segmentation DNN was trained using relatively good quality scans. To test the validity of predicting high vs low quality scans, we performed an ROC analysis on the held-out NNDSP dataset, where manual quality control ratings are available. We also did the same analysis using MRIQC (v0.10.5) \cite{esteban2017mriqc}, a recently published method that combines a wide range of automated QC algorithms. To statistically test whether any method significantly outperformed the other methods, we performed bootstrap sampling of the AUC for predicting scan quality from average uncertainty by sampling out-of-site test volumes. We performed 10,000 bootstrap samples, each with 418 volumes. The average ROC and AUC for the MAP, BD, SSD, and MRIQC methods are shown in Figure \ref{qc_roc}. The MAP, BD, and SSD networks all have significantly higher AUCs than MRIQC ($p=1.369e-4$,$p=1.272e-5$, and $p=1.381e-6$, respectively). Additionally, SSD had a significantly higher AUC than both MAP and BD ($p=1.156e-3$ and $p=1.042e-3$, respectively).

\section{Discussion}

Segmentation of structures in sMRI volumes is a critical pre-processing step in many neuroimaging analyses. However, these segmentations are currently generated using tools that can take a day or more for each subject \citep{Runtimes}, such as FreeSurfer. This computational cost can be prohibitive when scaling analyses up from hundreds to thousands of subjects. DNNs have recently been proposed to perform sMRI segmentation is seconds to minutes. In this paper, we developed a Bayesian DNN, using spike-and-slab dropout, with the goals of increasing the similarity of the DNN's predictions to the FreeSurfer segmentations and generating useful uncertainty estimates for these predictions.

In order to evaluate the proposed Bayesian network, we trained a standard deep neural network (DNN), using MAP estimation, to predict FreeSurfer segmentations from structural MRI (sMRI) volumes. We trained on a little under 10,000 sMRIs, obtained by combining approximately 70 different datasets (many of which, in turn, contain images from several sites, e.g. NKI, ABIDE, ADHD200). We used a separate test set of more than 1,000 sMRIs, drawn from the same datasets. The resulting standard DNN performs at the same level of state-of-the-art networks \citep{fedorov2017almost}. This result, however, was obtained by testing over an order of magnitude more test data, and many more sites, than those papers. We also tested performance on a completely separate dataset (NNDSP) from a site not encountered in training, which contained 418 sMRI volumes. Whereas Dice performance dropped slightly, this was less than what was observed in other studies \citep{roy2018quicknat,roy2018bayesian}; this suggests that we may be achieving better generalization by training on our larger and more diverse dataset, and we plan on testing this on more datasets from novel sites in the future. This is particularly important to us, as this network is meant to be used within an off-the-shelf tool\footnote{\url{ https://github.com/neuronets/nobrainer}}.

We demonstrated that the estimated uncertainty for the prediction at each voxel is a good indicator of whether the standard network makes an error in it, both in-site and out-of-site. The tool that produces the predicted segmentation volume for an input sMRI will also produce an uncertainty volume. We anticipate this being useful at various levels, e.g. to refine other tools that rely on segmentation images, or to to improve prediction models based on sMRI data (e.g. modification of calculation of cortical thickness, surface area, voxel selection or weighting in regression \citep{roy2018bayesian} or classification models, etc).

We also demonstrated that the average prediction uncertainty across voxels in the brain is an excellent indicator of  manual quality control ratings. Furthermore, it outperforms the best existing automated solution \citep{esteban2017mriqc}. Since automation is already used in large repositories (e.g. OpenMRI), we plan on offering our tool as an additional quality control measure.

Finally, we showed that a new Bayesian DNN using spike-and-slab dropout with learned model uncertainty was significantly better than previous approaches. This spike-and-slab method increased segmentation performance and improved the usefulness of output uncertainties compared both to a MAP DNN method and an MC Bernoulli dropout method, which has previously been used in the brain segmentation literature \citep{li2017compactness,roy2018bayesian}. These results show that Bayesian DNNs are a promising method for building brain segmentation and automated sMRI quality control tools. We have also made a version of ``Nobrainer", that incorporates the networks trained and evaluated in this paper, available for download and use within a Singularity/Docker container \footnote{\url{https://github.com/neuronets/kwyk}}.

We believe it may be possible to improve this segmentation processing, in that we did not use registration. One option would be to use various techniques for data augmentation (e.g. variation of image contrast, since that is pretty heterogeneous, rotations/translations of existing examples, addition of realistic noise, etc). Another would be to eliminate the need to divide the brain into sub-volumes, which loses some global information; this will become more feasible in GPUs with more memory. Finally, we plan on using post-processing of results (e.g. ensure some coherence between predictions for adjacent voxels, leverage off-the-shelf brain and tissue masking code).

\section*{Acknowledgments}

This research was supported (in part) by the Intramural Research Program of the NIMH (ZICMH002968). This work utilized the computational resources of the NIH HPC Biowulf cluster. (\url{http://hpc.nih.gov}). JK's and SG's contribution was supported by NIH R01 EB020740. 

\bibliographystyle{frontiersinSCNS_ENG_HUMS} 
\bibliography{test}

\begin{thebibliography}{60}
\providecommand{\natexlab}[1]{#1}
\expandafter\ifx\csname urlstyle\endcsname\relax
  \providecommand{\doi}[1]{doi:\discretionary{}{}{}#1}\else
  \providecommand{\doi}{doi:\discretionary{}{}{}\begingroup
  \urlstyle{rm}\Url}\fi
\providecommand{\selectlanguage}[1]{\relax}
\providecommand{\bibAnnoteFile}[1]{%
  \IfFileExists{#1}{\begin{quotation}\noindent\textsc{Key:} #1\\
  \textsc{Annotation:}\ \input{#1}\end{quotation}}{}}
\providecommand{\bibAnnote}[2]{%
  \begin{quotation}\noindent\textsc{Key:} #1\\
  \textsc{Annotation:}\ #2\end{quotation}}

\bibitem[{Abadi et~al.(2016)Abadi, Barham, Chen, Chen, Davis, Dean
  et~al.}]{abadi2016tensorflow}
Abadi, M., Barham, P., Chen, J., Chen, Z., Davis, A., Dean, J., et~al. (2016).
\newblock Tensorflow: A system for large-scale machine learning.
\newblock In \emph{12th $\{$USENIX$\}$ Symposium on Operating Systems Design
  and Implementation ($\{$OSDI$\}$ 16)}. 265--283
\bibAnnoteFile{abadi2016tensorflow}

\bibitem[{Alexander et~al.(2017)Alexander, Escalera, Ai, Andreotti, Febre,
  Mangone et~al.}]{alexander2017open}
Alexander, L.~M., Escalera, J., Ai, L., Andreotti, C., Febre, K., Mangone, A.,
  et~al. (2017).
\newblock An open resource for transdiagnostic research in pediatric mental
  health and learning disorders.
\newblock \emph{Scientific data} 4, 170181
\bibAnnoteFile{alexander2017open}

\bibitem[{Bellec et~al.(2017)Bellec, Chu, Chouinard-Decorte, Benhajali,
  Margulies, and Craddock}]{bellec2017neuro}
Bellec, P., Chu, C., Chouinard-Decorte, F., Benhajali, Y., Margulies, D.~S.,
  and Craddock, R.~C. (2017).
\newblock The neuro bureau adhd-200 preprocessed repository.
\newblock \emph{Neuroimage} 144, 275--286
\bibAnnoteFile{bellec2017neuro}

\bibitem[{Biswal et~al.(2010)Biswal, Mennes, Zuo, Gohel, Kelly, Smith
  et~al.}]{biswal2010toward}
Biswal, B.~B., Mennes, M., Zuo, X.-N., Gohel, S., Kelly, C., Smith, S.~M.,
  et~al. (2010).
\newblock Toward discovery science of human brain function.
\newblock \emph{Proceedings of the National Academy of Sciences} 107,
  4734--4739
\bibAnnoteFile{biswal2010toward}

\bibitem[{Blumenthal et~al.(2002)Blumenthal, Zijdenbos, Molloy, and
  Giedd}]{blumenthal2002motion}
Blumenthal, J.~D., Zijdenbos, A., Molloy, E., and Giedd, J.~N. (2002).
\newblock Motion artifact in magnetic resonance imaging: implications for
  automated analysis.
\newblock \emph{Neuroimage} 16, 89--92
\bibAnnoteFile{blumenthal2002motion}

\bibitem[{Blundell et~al.(2015)Blundell, Cornebise, Kavukcuoglu, and
  Wierstra}]{blundell2015weight}
Blundell, C., Cornebise, J., Kavukcuoglu, K., and Wierstra, D. (2015).
\newblock Weight uncertainty in neural networks.
\newblock In \emph{International Conference on Machine Learning}. 1613--1622
\bibAnnoteFile{blundell2015weight}

\bibitem[{Cardoso et~al.(2015)Cardoso, Modat, Wolz, Melbourne, Cash, Rueckert
  et~al.}]{cardoso2015geodesic}
Cardoso, M.~J., Modat, M., Wolz, R., Melbourne, A., Cash, D., Rueckert, D.,
  et~al. (2015).
\newblock Geodesic information flows: spatially-variant graphs and their
  application to segmentation and fusion.
\newblock \emph{IEEE transactions on medical imaging} 34, 1976--1988
\bibAnnoteFile{cardoso2015geodesic}

\bibitem[{Der~Kiureghian and Ditlevsen(2009)}]{der2009aleatory}
Der~Kiureghian, A. and Ditlevsen, O. (2009).
\newblock Aleatory or epistemic? does it matter?
\newblock \emph{Structural Safety} 31, 105--112
\bibAnnoteFile{der2009aleatory}

\bibitem[{Di~Martino et~al.(2014)Di~Martino, Yan, Li, Denio, Castellanos,
  Alaerts et~al.}]{di2014autism}
Di~Martino, A., Yan, C.-G., Li, Q., Denio, E., Castellanos, F.~X., Alaerts, K.,
  et~al. (2014).
\newblock The autism brain imaging data exchange: Towards a large-scale
  evaluation of the intrinsic brain architecture in autism.
\newblock \emph{Molecular Psychiatry} 19, 659
\bibAnnoteFile{di2014autism}

\bibitem[{di~Oleggio~Castello et~al.(2017)di~Oleggio~Castello, Halchenko,
  Guntupalli, Gors, and Gobbini}]{di2017neural}
di~Oleggio~Castello, M.~V., Halchenko, Y.~O., Guntupalli, J.~S., Gors, J.~D.,
  and Gobbini, M.~I. (2017).
\newblock The neural representation of personally familiar and unfamiliar faces
  in the distributed system for face perception.
\newblock \emph{Scientific Reports} 7
\bibAnnoteFile{di2017neural}

\bibitem[{Dolz et~al.(2018)Dolz, Desrosiers, and Ayed}]{dolz20183d}
Dolz, J., Desrosiers, C., and Ayed, I.~B. (2018).
\newblock 3d fully convolutional networks for subcortical segmentation in mri:
  A large-scale study.
\newblock \emph{NeuroImage} 170, 456--470
\bibAnnoteFile{dolz20183d}

\bibitem[{Esteban et~al.(2017)Esteban, Birman, Schaer, Koyejo, Poldrack, and
  Gorgolewski}]{esteban2017mriqc}
Esteban, O., Birman, D., Schaer, M., Koyejo, O.~O., Poldrack, R.~A., and
  Gorgolewski, K.~J. (2017).
\newblock Mriqc: Advancing the automatic prediction of image quality in mri
  from unseen sites.
\newblock \emph{PloS one} 12, e0184661
\bibAnnoteFile{esteban2017mriqc}

\bibitem[{Fedorov et~al.(2017{\natexlab{a}})Fedorov, Damaraju, Calhoun, and
  Plis}]{fedorov2017almost}
Fedorov, A., Damaraju, E., Calhoun, V., and Plis, S. (2017{\natexlab{a}}).
\newblock Almost instant brain atlas segmentation for large-scale studies.
\newblock \emph{arXiv preprint arXiv:1711.00457}
\bibAnnoteFile{fedorov2017almost}

\bibitem[{Fedorov et~al.(2017{\natexlab{b}})Fedorov, Johnson, Damaraju, Ozerin,
  Calhoun, and Plis}]{fedorov2017end}
Fedorov, A., Johnson, J., Damaraju, E., Ozerin, A., Calhoun, V., and Plis, S.
  (2017{\natexlab{b}}).
\newblock End-to-end learning of brain tissue segmentation from imperfect
  labeling.
\newblock In \emph{International Joint Conference on Neural Networks} (IEEE),
  3785--3792
\bibAnnoteFile{fedorov2017end}

\bibitem[{Fischl(2012)}]{fischl2012FreeSurfer}
Fischl, B. (2012).
\newblock Freesurfer.
\newblock \emph{Neuroimage} 62
\bibAnnoteFile{fischl2012FreeSurfer}

\bibitem[{FreeSurfer(2018)}]{Runtimes}
FreeSurfer (2018).
\newblock {}reconallruntimes Accessed: 2018-12-04
\bibAnnoteFile{Runtimes}

\bibitem[{Gal(2016)}]{gal2016uncertainty}
Gal, Y. (2016).
\newblock \emph{Uncertainty in Deep Learning}.
\newblock Ph.D. thesis, University of Cambridge
\bibAnnoteFile{gal2016uncertainty}

\bibitem[{Gal and Ghahramani(2016)}]{gal2016dropout}
Gal, Y. and Ghahramani, Z. (2016).
\newblock Dropout as a {Bayesian} approximation: Representing model uncertainty
  in deep learning.
\newblock In \emph{International Conference on Machine Learning}. 1050--1059
\bibAnnoteFile{gal2016dropout}

\bibitem[{Gal et~al.(2017)Gal, Hron, and Kendall}]{gal2017concrete}
Gal, Y., Hron, J., and Kendall, A. (2017).
\newblock Concrete dropout.
\newblock In \emph{Advances in Neural Information Processing Systems}.
  3581--3590
\bibAnnoteFile{gal2017concrete}

\bibitem[{Graves(2011)}]{graves2011practical}
Graves, A. (2011).
\newblock Practical variational inference for neural networks.
\newblock In \emph{Advances in Neural Information Processing Systems}.
  2348--2356
\bibAnnoteFile{graves2011practical}

\bibitem[{Guo et~al.(2017)Guo, Pleiss, Sun, and
  Weinberger}]{guo2017calibration}
Guo, C., Pleiss, G., Sun, Y., and Weinberger, K.~Q. (2017).
\newblock On calibration of modern neural networks.
\newblock In \emph{Proceedings of the 34th International Conference on Machine
  Learning-Volume 70}. 1321--1330
\bibAnnoteFile{guo2017calibration}

\bibitem[{Hastie et~al.(2005)Hastie, Tibshirani, Friedman, and
  Franklin}]{hastie2005elements}
Hastie, T., Tibshirani, R., Friedman, J., and Franklin, J. (2005).
\newblock The elements of statistical learning: data mining, inference and
  prediction.
\newblock \emph{The Mathematical Intelligencer} 27, 83--85
\bibAnnoteFile{hastie2005elements}

\bibitem[{Haxby et~al.(2011)Haxby, Guntupalli, Connolly, Halchenko, Conroy,
  Gobbini et~al.}]{haxby2011common}
Haxby, J.~V., Guntupalli, J.~S., Connolly, A.~C., Halchenko, Y.~O., Conroy,
  B.~R., Gobbini, M.~I., et~al. (2011).
\newblock A common, high-dimensional model of the representational space in
  human ventral temporal cortex.
\newblock \emph{Neuron} 72, 404--416
\bibAnnoteFile{haxby2011common}

\bibitem[{Hinton and Van~Camp(1993)}]{hinton1993keeping}
Hinton, G.~E. and Van~Camp, D. (1993).
\newblock Keeping the neural networks simple by minimizing the description
  length of the weights.
\newblock In \emph{Proceedings of the sixth annual conference on Computational
  learning theory} (ACM), 5--13
\bibAnnoteFile{hinton1993keeping}

\bibitem[{Holmes et~al.(2015)Holmes, Hollinshead, O’Keefe, Petrov, Fariello,
  Wald et~al.}]{holmes2015brain}
Holmes, A.~J., Hollinshead, M.~O., O’Keefe, T.~M., Petrov, V.~I., Fariello,
  G.~R., Wald, L.~L., et~al. (2015).
\newblock Brain genomics superstruct project initial data release with
  structural, functional, and behavioral measures.
\newblock \emph{Scientific data} 2, 150031
\bibAnnoteFile{holmes2015brain}

\bibitem[{Hron et~al.(2018)Hron, Matthews, and
  Ghahramani}]{hron2018variational}
Hron, J., Matthews, A.~G., and Ghahramani, Z. (2018).
\newblock Variational bayesian dropout: pitfalls and fixes.
\newblock In \emph{International Conference on Machine Learning}. 2024--2033
\bibAnnoteFile{hron2018variational}

\bibitem[{Keator et~al.(2016)Keator, van Erp, Turner, Glover, Mueller, Liu
  et~al.}]{keator2016function}
Keator, D.~B., van Erp, T.~G., Turner, J.~A., Glover, G.~H., Mueller, B.~A.,
  Liu, T.~T., et~al. (2016).
\newblock The function biomedical informatics research network data repository.
\newblock \emph{Neuroimage} 124, 1074--1079
\bibAnnoteFile{keator2016function}

\bibitem[{Kendall and Gal(2017)}]{kendall2017uncertainties}
Kendall, A. and Gal, Y. (2017).
\newblock What uncertainties do we need in bayesian deep learning for computer
  vision?
\newblock In \emph{Advances in neural information processing systems}.
  5574--5584
\bibAnnoteFile{kendall2017uncertainties}

\bibitem[{Kennedy et~al.(2012)Kennedy, Haselgrove, Hodge, Rane, Makris, and
  Frazier}]{kennedy2012candishare}
[Dataset] Kennedy, D.~N., Haselgrove, C., Hodge, S.~M., Rane, P.~S., Makris,
  N., and Frazier, J.~A. (2012).
\newblock Candishare: a resource for pediatric neuroimaging data
\bibAnnoteFile{kennedy2012candishare}

\bibitem[{Keshavan et~al.(2018)Keshavan, Yeatman, and Rokem}]{Keshavan363382}
Keshavan, A., Yeatman, J., and Rokem, A. (2018).
\newblock Combining citizen science and deep learning to amplify expertise in
  neuroimaging.
\newblock \emph{bioRxiv} \doi{10.1101/363382}
\bibAnnoteFile{Keshavan363382}

\bibitem[{Kingma and Ba(2015)}]{kingma2014adam}
Kingma, D.~P. and Ba, J. (2015).
\newblock Adam: A method for stochastic optimization.
\newblock In \emph{International Conference on Learning Representations}
\bibAnnoteFile{kingma2014adam}

\bibitem[{Kingma et~al.(2015)Kingma, Salimans, and
  Welling}]{kingma2015variational}
Kingma, D.~P., Salimans, T., and Welling, M. (2015).
\newblock Variational dropout and the local reparameterization trick.
\newblock In \emph{Advances in Neural Information Processing Systems}.
  2575--2583
\bibAnnoteFile{kingma2015variational}

\bibitem[{Lee et~al.(2018)Lee, Gustavo, Migineishvili, Nielson, Bandettini,
  Shaw et~al.}]{lee2018automated}
Lee, J.~A., Gustavo, S., Migineishvili, N., Nielson, D., Bandettini, P.~A.,
  Shaw, P., et~al. (2018).
\newblock Automated quality control methods applied to a novel dataset.
\newblock \doi{https://doi.org/10.15154/1463004}.
\newblock Organization for Human Brain Mapping (OHBM) Annual Meeting
\bibAnnoteFile{lee2018automated}

\bibitem[{Li et~al.(2017)Li, Wang, Fidon, Ourselin, Cardoso, and
  Vercauteren}]{li2017compactness}
Li, W., Wang, G., Fidon, L., Ourselin, S., Cardoso, M.~J., and Vercauteren, T.
  (2017).
\newblock On the compactness, efficiency, and representation of {3D}
  convolutional networks: Brain parcellation as a pretext task.
\newblock In \emph{International Conference on Information Processing in
  Medical Imaging} (Springer), 348--360
\bibAnnoteFile{li2017compactness}

\bibitem[{Liu et~al.(2017)Liu, Wei, Chen, Yang, Meng, Wu
  et~al.}]{liu2017longitudinal}
Liu, W., Wei, D., Chen, Q., Yang, W., Meng, J., Wu, G., et~al. (2017).
\newblock Longitudinal test-retest neuroimaging data from healthy young adults
  in southwest china.
\newblock \emph{Scientific data} 4, 170017
\bibAnnoteFile{liu2017longitudinal}

\bibitem[{Louizos and Welling(2017)}]{louizos2017multiplicative}
Louizos, C. and Welling, M. (2017).
\newblock Multiplicative normalizing flows for variational {Bayesian} neural
  networks.
\newblock In \emph{International Conference on Machine Learning}. 2218--2227
\bibAnnoteFile{louizos2017multiplicative}

\bibitem[{Marcus et~al.(2007)Marcus, Wang, Parker, Csernansky, Morris, and
  Buckner}]{marcus2007open}
Marcus, D.~S., Wang, T.~H., Parker, J., Csernansky, J.~G., Morris, J.~C., and
  Buckner, R.~L. (2007).
\newblock Open access series of imaging studies (oasis): cross-sectional mri
  data in young, middle aged, nondemented, and demented older adults.
\newblock \emph{Journal of cognitive neuroscience} 19, 1498--1507
\bibAnnoteFile{marcus2007open}

\bibitem[{Mazziotta et~al.(2001)Mazziotta, Toga, Evans, Fox, Lancaster, Zilles
  et~al.}]{mazziotta2001probabilistic}
Mazziotta, J., Toga, A., Evans, A., Fox, P., Lancaster, J., Zilles, K., et~al.
  (2001).
\newblock A probabilistic atlas and reference system for the human brain:
  International consortium for brain mapping (icbm).
\newblock \emph{Philosophical Transactions of the Royal Society of London B:
  Biological Sciences} 356, 1293--1322
\bibAnnoteFile{mazziotta2001probabilistic}

\bibitem[{McClure and Kriegeskorte(2017)}]{mcclure2017representing}
McClure, P. and Kriegeskorte, N. (2017).
\newblock Robustly representing uncertainty in deep neural networks through
  sampling.
\newblock In \emph{NIPS Bayesian Deep Learning Workshop}
\bibAnnoteFile{mcclure2017representing}

\bibitem[{McClure et~al.(2018)McClure, Zheng, Kaczmarzyk, Rogers-Lee, Ghosh,
  Nielson et~al.}]{mcclure2018distributed}
McClure, P., Zheng, C.~Y., Kaczmarzyk, J., Rogers-Lee, J., Ghosh, S., Nielson,
  D., et~al. (2018).
\newblock Distributed weight consolidation: A brain segmentation case study.
\newblock In \emph{Advances in Neural Information Processing Systems 31}, eds.
  S.~Bengio, H.~Wallach, H.~Larochelle, K.~Grauman, N.~Cesa-Bianchi, and
  R.~Garnett. 4093--4103
\bibAnnoteFile{mcclure2018distributed}

\bibitem[{Molchanov et~al.(2017)Molchanov, Ashukha, and
  Vetrov}]{molchanov2017variational}
Molchanov, D., Ashukha, A., and Vetrov, D. (2017).
\newblock Variational dropout sparsifies deep neural networks.
\newblock In \emph{International Conference on Machine Learning}. 2498--2507
\bibAnnoteFile{molchanov2017variational}

\bibitem[{Mueller et~al.(2005)Mueller, Weiner, Thal, Petersen, Jack, Jagust
  et~al.}]{mueller2005alzheimer}
Mueller, S.~G., Weiner, M.~W., Thal, L.~J., Petersen, R.~C., Jack, C., Jagust,
  W., et~al. (2005).
\newblock The alzheimer's disease neuroimaging initiative.
\newblock \emph{Neuroimaging Clinics} 15, 869--877
\bibAnnoteFile{mueller2005alzheimer}

\bibitem[{Nastase et~al.(2017)Nastase, Connolly, Oosterhof, Halchenko,
  Guntupalli, Visconti~di Oleggio~Castello et~al.}]{nastase2017attention}
Nastase, S.~A., Connolly, A.~C., Oosterhof, N.~N., Halchenko, Y.~O.,
  Guntupalli, J.~S., Visconti~di Oleggio~Castello, M., et~al. (2017).
\newblock Attention selectively reshapes the geometry of distributed semantic
  representation.
\newblock \emph{Cerebral Cortex} 27, 4277--4291
\bibAnnoteFile{nastase2017attention}

\bibitem[{Nguyen et~al.(2018)Nguyen, Li, Bui, and
  Turner}]{nguyen2018variational}
Nguyen, C.~V., Li, Y., Bui, T.~D., and Turner, R.~E. (2018).
\newblock Variational continual learning.
\newblock In \emph{International Conference on Learning Representations}
\bibAnnoteFile{nguyen2018variational}

\bibitem[{Nooner et~al.(2012)Nooner, Colcombe, Tobe, Mennes, Benedict, Moreno
  et~al.}]{nooner2012nki}
Nooner, K.~B., Colcombe, S., Tobe, R., Mennes, M., Benedict, M., Moreno, A.,
  et~al. (2012).
\newblock The {NKI-Rockland} sample: A model for accelerating the pace of
  discovery science in psychiatry.
\newblock \emph{Frontiers in Neuroscience} 6, 152
\bibAnnoteFile{nooner2012nki}

\bibitem[{O’Connor et~al.(2017)O’Connor, Potler, Kovacs, Xu, Ai, Pellman
  et~al.}]{o2017healthy}
O’Connor, D., Potler, N.~V., Kovacs, M., Xu, T., Ai, L., Pellman, J., et~al.
  (2017).
\newblock The healthy brain network serial scanning initiative: a resource for
  evaluating inter-individual differences and their reliabilities across scan
  conditions and sessions.
\newblock \emph{Gigascience} 6, 1--14
\bibAnnoteFile{o2017healthy}

\bibitem[{Petersen et~al.(2010)Petersen, Aisen, Beckett, Donohue, Gamst, Harvey
  et~al.}]{petersen2010alzheimer}
Petersen, R.~C., Aisen, P., Beckett, L., Donohue, M., Gamst, A., Harvey, D.,
  et~al. (2010).
\newblock Alzheimer's disease neuroimaging initiative (adni): clinical
  characterization.
\newblock \emph{Neurology} 74, 201--209
\bibAnnoteFile{petersen2010alzheimer}

\bibitem[{Poldrack et~al.(2013)Poldrack, Barch, Mitchell, Wager, Wagner, Devlin
  et~al.}]{poldrack2013toward}
Poldrack, R.~A., Barch, D.~M., Mitchell, J., Wager, T., Wagner, A.~D., Devlin,
  J.~T., et~al. (2013).
\newblock Toward open sharing of task-based fmri data: the openfmri project.
\newblock \emph{Frontiers in neuroinformatics} 7, 12
\bibAnnoteFile{poldrack2013toward}

\bibitem[{Robbins and Monro(1951)}]{robbins1951stochastic}
Robbins, H. and Monro, S. (1951).
\newblock A stochastic approximation method.
\newblock \emph{The Annals of Mathematical Statistics} , 400--407
\bibAnnoteFile{robbins1951stochastic}

\bibitem[{Rohlfing(2012)}]{rohlfing2012image}
Rohlfing, T. (2012).
\newblock Image similarity and tissue overlaps as surrogates for image
  registration accuracy: widely used but unreliable.
\newblock \emph{IEEE transactions on medical imaging} 31, 153--163
\bibAnnoteFile{rohlfing2012image}

\bibitem[{Ronneberger et~al.(2015)Ronneberger, Fischer, and
  Brox}]{ronneberger2015u}
Ronneberger, O., Fischer, P., and Brox, T. (2015).
\newblock U-net: Convolutional networks for biomedical image segmentation.
\newblock In \emph{International Conference on Medical Image Computing and
  Computer-Assisted Intervention} (Springer), 234--241
\bibAnnoteFile{ronneberger2015u}

\bibitem[{Roy et~al.(2018{\natexlab{a}})Roy, Conjeti, Navab, and
  Wachinger}]{roy2018bayesian}
Roy, A.~G., Conjeti, S., Navab, N., and Wachinger, C. (2018{\natexlab{a}}).
\newblock Bayesian quicknat: Model uncertainty in deep whole-brain segmentation
  for structure-wise quality control.
\newblock \emph{arXiv preprint arXiv:1811.09800}
\bibAnnoteFile{roy2018bayesian}

\bibitem[{Roy et~al.(2018{\natexlab{b}})Roy, Conjeti, Navab, and
  Wachinger}]{roy2018quicknat}
Roy, A.~G., Conjeti, S., Navab, N., and Wachinger, C. (2018{\natexlab{b}}).
\newblock {QuickNAT}: Segmenting {MRI} neuroanatomy in 20 seconds.
\newblock \emph{arXiv preprint arXiv:1801.04161}
\bibAnnoteFile{roy2018quicknat}

\bibitem[{Srivastava et~al.(2014)Srivastava, Hinton, Krizhevsky, Sutskever, and
  Salakhutdinov}]{srivastava2014dropout}
Srivastava, N., Hinton, G., Krizhevsky, A., Sutskever, I., and Salakhutdinov,
  R. (2014).
\newblock Dropout: a simple way to prevent neural networks from overfitting.
\newblock \emph{The Journal of Machine Learning Research} 15, 1929--1958
\bibAnnoteFile{srivastava2014dropout}

\bibitem[{Titsias and L{\'a}zaro-Gredilla(2011)}]{titsias2011spike}
Titsias, M.~K. and L{\'a}zaro-Gredilla, M. (2011).
\newblock Spike and slab variational inference for multi-task and multiple
  kernel learning.
\newblock In \emph{Advances in neural information processing systems}.
  2339--2347
\bibAnnoteFile{titsias2011spike}

\bibitem[{Van~Essen et~al.(2013)Van~Essen, Smith, Barch, Behrens, Yacoub,
  Ugurbil et~al.}]{van2013wu}
Van~Essen, D.~C., Smith, S.~M., Barch, D.~M., Behrens, T.~E., Yacoub, E.,
  Ugurbil, K., et~al. (2013).
\newblock The {WU-Minn} human connectome project: An overview.
\newblock \emph{NeuroImage} 80, 62--79
\bibAnnoteFile{van2013wu}

\bibitem[{Vázquez et~al.(2016)Vázquez, Whitfield-Gabrieli, Bauer, Barrios,
  and A}]{barrios}
Vázquez, P.~G., Whitfield-Gabrieli, S., Bauer, C.~C., Barrios, and A, F.
  (2016).
\newblock Brain functional connectivity of hypnosis without target suggestion.
  an intrinsic hypnosis rs-fmri study.
\newblock \emph{(under review)}
\bibAnnoteFile{barrios}

\bibitem[{Wei et~al.(2018)Wei, Zhuang, Chen, Yang, Liu, Wang
  et~al.}]{wei2018structural}
Wei, D., Zhuang, K., Chen, Q., Yang, W., Liu, W., Wang, K., et~al. (2018).
\newblock Structural and functional mri from a cross-sectional southwest
  university adult lifespan dataset (sald).
\newblock \emph{bioRxiv} , 177279
\bibAnnoteFile{wei2018structural}

\bibitem[{Yu and Koltun(2015)}]{yu2015multi}
Yu, F. and Koltun, V. (2015).
\newblock Multi-scale context aggregation by dilated convolutions.
\newblock In \emph{International Conference on Learning Representations}
\bibAnnoteFile{yu2015multi}

\bibitem[{Zuo et~al.(2014)Zuo, Anderson, Bellec, Birn, Biswal, Blautzik
  et~al.}]{zuo2014open}
Zuo, X.-N., Anderson, J.~S., Bellec, P., Birn, R.~M., Biswal, B.~B., Blautzik,
  J., et~al. (2014).
\newblock An open science resource for establishing reliability and
  reproducibility in functional connectomics.
\newblock \emph{Scientific data} 1, 140049
\bibAnnoteFile{zuo2014open}

\end{thebibliography}


\end{document}